\documentclass[11pt]{article}

\usepackage[final]{acl}
\usepackage{times}
\usepackage{latexsym}
\usepackage[T1]{fontenc}
\usepackage{microtype}
\usepackage{inconsolata}
\usepackage{graphicx}
\usepackage{booktabs}
\usepackage{amsmath}
\usepackage{url}

\title{Looking Complete Is Not Being Grounded: \\ Evaluating LLM Sitreps Against Human-Authored Situation Reports}

\author{Sachin Kumar\thanks{\ \ This research was conducted independently and does not reflect the views or represent work done at LexisNexis.} \\
  LexisNexis, USA \\
  \texttt{sachinkumar.ait@live.com}}

\begin{document}
\maketitle

\begin{abstract}
Evaluation for automatic crisis reporting should measure what practitioners need, not what is convenient to compare. Through LLM-assisted, author-approved content analysis, we derive a 15-slot information schema from professional humanitarian situation reports using qualitative content analysis, then evaluate 121 human and 261 LLM-generated reports across five disasters and three hazard types. Under \texttt{google/gemini-3.7-flash} on Cyclone Idai, the best machine arm is nominally above the empirical human ceiling (0.827 vs. 0.821), but the difference is statistically indistinguishable from zero (95\% bootstrap CI $[-.049,.059]$; permutation $p=.852$), and the ordering does not replicate under \texttt{openai/gpt-5.6-luna}. On a blind 100-item Idai sample, the author agrees with the Gemini judge at 0.910 accuracy and unweighted Cohen's $\kappa=.799$. All four judges clear the plan's raw-agreement criterion of .7; only Gemini reaches .7 on chance-corrected $\kappa$. Across the three primary events, machine reports reproduce 5.4--11.0\% of same-day official figures. Across those same events, raw exact-number matching links 14.3--19.6\% of official figures and 85.4--90.1\% of machine figures to the social stream; manual audit confirms 45/50 declared absences, 4/25 official matches, and 24/25 machine matches semantically. The issuing-provenance gap is positive in all 12 observed event--judge cells: report date, organisation, series, and authority were not supplied at professional-report coverage in any cell. On the three primary events, ROUGE-L is uninformative, BERTScore is negatively associated with completeness after multiplicity correction, and a generic LLM judge changes sign.
\end{abstract}

\section{Introduction}

A situation report (sitrep) is not simply a summary. It is a dated, attributed operational document that supports the coordination of a response by reporting current needs, actions, gaps, and key figures \citep{unocha-sitrep}. The distinction matters for the humanitarian teams who would act on such reports and for the LT4CPR shared task \citep{lt4cpr-2026-shared-task}: a fluent report can mention every expected topic while omitting the quantitative and provenance-bearing information on which operational use depends.

Most automatic evaluation begins with available instruments: overlap with a reference, embedding similarity, or a general-purpose model asked whether an output is good. We instead begin with practice. We ask what information professional humanitarian organisations actually place in sitreps, how much of it a system can recover from social media, and whether common metrics recognise that recovery.

We make three contributions. First, we induce a 15-slot schema from professional reports through LLM-assisted qualitative content analysis, and report what each computational step contributed and what it excluded. Second, we define \emph{operational completeness} relative to an empirical human ceiling and locate a \emph{visibility ceiling}: information that is present in official reports but absent from the source stream. Third, we meta-evaluate ROUGE-L, BERTScore, and a generic LLM judge against operational completeness. The apparent paradox is the paper's central result: machine sitreps look complete by slot presence, but their quantitative picture is bounded by an input stream containing only a small fraction of the official record. We close with implications for humanitarian practice (Section~\ref{sec:practice}) and for shared-task evaluation (Section~\ref{sec:shared-task}).

\section{Related Work}

Crisis informatics has long treated social media as a rapid source for situational awareness, while also documenting its noise, brevity, and uneven coverage \citep{imran-etal-2016-twitter,alam-etal-2021-humaid}. Recent work surveys the resulting language-technology ecosystem and its operational challenges \citep{d-lewis-etal-2025-tapping}. Our question differs from message classification: we compare a generated operational product against reporting practice and explicitly measure what the input medium did not supply in this corpus.

Crisis summarisation is the nearest task family. CrisisLTLSum extracts and abstractively summarises 1,000 local Twitter timelines \citep{rajaby-faghihi-etal-2022-crisisltlsum}; CrisisFACTS produces event-day timelines from social and news streams and evaluates pooled facts for comprehensiveness and redundancy \citep{buntain-etal-2023-crisisfacts}; prompt-based LLM systems have also aggregated geolocated social feedback into disaster reports \citep{cantini-etal-2025-reporting}. The current LT4CPR task maps crisis posts to a structured official template \citep{lt4cpr-2026-shared-task}.We add a combination these benchmarks do not provide: a slot-level comparison of model reports with independently published humanitarian sitreps, a practice-derived schema, and an audit of which official information was visible in the input. CrisisFACTS's earlier events used official ICS-209 incident forms as an alternative ground truth \citep{buntain-etal-2023-crisisfacts}, but without a practice-derived schema or an input-visibility audit.

Our schema also differs from NLP event-schema induction, which learns entity and event-role structures from corpora \citep{chambers-2013-event}, and from methods that optimise prompt tokens for a downstream objective \citep{shin-etal-2020-autoprompt}. It is an LLM-assisted, author-approved content-analysis description of recurrent document information types. Operational completeness inherits the content-unit intuition of Pyramid and later Summary Content Unit evaluation \citep{nenkova-passonneau-2004-evaluating,nawrath-etal-2024-role}, but its units are practice expectations induced across reports, its categories allow partial credit, and its reference point is an empirical human ceiling.

Summarisation evaluation commonly uses lexical overlap such as ROUGE \citep{lin-2004-rouge} and contextual similarity such as BERTScore \citep{zhang-etal-2020-bertscore}. Prior analyses show that such metrics can track topical similarity rather than information quality and that realistic system-level ROUGE correlations may approach zero \citep{deutsch-roth-2021-information,deutsch-etal-2022-examining}. The latter critique is system-level, whereas our main Table~\ref{tab:corr} correlations are between reports; we therefore also report a descriptive ordering of the four generator--prompt arms. LLM judges offer flexible reference-free evaluation \citep{liu-etal-2023-geval}, but can favour their own generations \citep{panickssery-etal-2024-llm}. We test all three metric families against a domain-derived criterion rather than general ratings.

\section{Data}

\begin{table*}[t]
\centering
\small
\setlength{\tabcolsep}{6pt}
\begin{tabular}{lrrrr}
\toprule
Event & Human coll./eval. & Machine gen./eval. & Judges & Judged pairs \\
\midrule
Cyclone Idai 2019 & 59 / 56 & 104 / 102 & 4 & 9,302 \\
Hurricane Irma 2017 & 19 / 19 & 48 / 48 & 2 & 2,010 \\
Hurricane Maria 2017 & 25 / 25 & 44 / 43 & 2 & 2,040 \\
Puebla earthquake 2017 & 15 / 14 & 48 / 48 & 2 & 1,856 \\
Kerala floods 2018 & 7 / 7 & 20 / 20 & 2 & 808 \\
\midrule
Total & 125 / 121 & 264 / 261 & -- & 16,016 \\
\bottomrule
\end{tabular}
\caption{Manifest-reconciled corpus. A ``judged pair'' is one parsable verdict for one (judge, physical document, slot); repeated judges multiply the count. Each cell is $\sum_j\sum_d|\mathcal{O}_j(d)|$, and the total equals the column sum. Idai uses DeepSeek, Luna, Gemini, and Qwen; other events use Luna and Gemini. Human and machine exclusions are shown uniformly as collected/generated versus evaluated.}
\label{tab:corpus}
\end{table*}

\paragraph{Professional reports.}
We manually collected public ReliefWeb products from OCHA, IFRC, UNICEF, WHO, CDEMA, and other humanitarian organisations. ReliefWeb's API required a pre-approved \texttt{appname} that had not been granted, while scripted site requests returned an HTTP 202 bot challenge. Manual download followed by automated scanning and bookkeeping was therefore the compliant collection route. The scanner extracted dates, organisations, and titles; content hashes removed duplicates. Table~\ref{tab:corpus} distinguishes collection from evaluation: three Idai rows are exact SHA-256 duplicates. Puebla's fifteenth PDF ingested to \texttt{\_2.json}; the manifest reader uniformly reserves leading-underscore files for helpers, so it was excluded, reconciling 15 collected versus 14 evaluated. Relative to $15J(H_{\rm eval}+M_{\rm eval})$, unusable judgements number 178/0/0/4/2 by table row; subtracting them produces every last-column cell. The supplement records titles, sources, verified dates, exact-title ReliefWeb locators, hashes, and the exclusion rule for all 125 products. We do not assume every ReliefWeb product is the same genre: most of Puebla's evaluated products are infographics or snapshots (10 of 14), so its score is not used as a sitrep ceiling, and Idai's 56 include 19 flash updates, three snapshots, and four IFRC DREF or emergency-appeal documents.

\paragraph{Social streams and machine reports.}
We use the English-language HumAID corpus \citep{alam-etal-2021-humaid}, which contains tweet identifiers and text across multiple disasters. We derive UTC dates from identifier snowflakes, bucket posts by event-day, and generate on days with at least 30 posts (for Kerala, only the first five such days, to stay within our cost limit); each generation reads a seed-17 sample of at most 400 of that day's posts, stratified by HumAID label, while source availability is measured against the complete day. We cross \texttt{api-strong}=\texttt{deepseek-v4-pro} and \texttt{api-fast}=\texttt{deepseek-v4-flash} with generic and schema-guided prompts. Calls ran 30--31 August 2026 UTC. The public aliases were then the Pro GA update dated 13 August and the Flash-0731 update dated 31 July; DeepSeek exposed no immutable snapshot suffix \citep{deepseek-2026-changelog}. The schema-guided prompt supplies the 15 slot names and definitions and asks for the names as section headings. Two empty Idai generations and one empty Maria generation are excluded from every judge and metric. Social-media text is neither redistributed nor quoted in this paper.

\section{Inducing a Practice-Derived Schema}
\label{sec:schema}

\paragraph{Interdisciplinary methods note.}
We borrow the phases of inductive content analysis---preparing material, organising codes into categories, and reporting an auditable structure \citep{elo-kyngas-2008,krippendorff-2018}---but a model, not the researcher, did the open coding. Qualitative researchers interviewed by \citet{schroeder-etal-2025-llms} worried that model-led coding imposes the model's interpretations and changes what researchers notice; \citeauthor{schroeder-etal-2025-llms} recommend validating each LLM task, for example against human coding, as \citet{wang-etal-2025-lata} did by comparing LLM-generated codes and themes with a prior human analysis of the same data. We have no human coding of our reports to compare against. We therefore separate computation that organised material (embedding and clustering near-synonymous codes) from steps that structured the conceptual space: our coding instructions and the model's codes set the vocabulary, a 25-cluster cap chose what reached review, and model-written definitions and an AI-drafted merge, which the author approved, set category boundaries. We treat the schema as one defensible description of Idai reporting practice, not the only one.

\paragraph{Open coding.}
The induction set is all 56 evaluated Idai reports (OCHA 44, IFRC 4, UNICEF 3, WFP 2, WHO 2, IOM 1): a sampler stratified by organisation and response phase (thirds of the collection's date range) requested 60, so every available report was included, and most are early-response products (45 early, eight middle, two late, one undated when sampled). \texttt{deepseek-v4-pro} open-coded each report in a separate call, instructed to record only information types present in it, each with an internal evidence span and location. The instructions also named five reporting conventions to record (attribution, as-of dates, hedged figures, revision language, and figures with sources) and gave four example code names, three of which (casualty figures, access constraints, funding-appeal status) became the most frequent code names of three candidates, along with four example section headings for locating codes. The conventions and example names echo our project plan, so these categories' prominence reflects our instructions as well as the reports. Reports longer than 30,000 characters (12 of 56, including all four from IFRC) were coded from their first 24,000 and last 6,000 characters, so 13\% of the induction text was never coded. The pass produced 1,360 codes; evidence spans remain in the private audit trail and are not reproduced here.

\paragraph{Consolidation.}
We embedded code names and definitions with MiniLM sentence embeddings and clustered them (average linkage, cosine distance, threshold .35), yielding 382 clusters. Clusters were ranked by the number of reports they covered, then by code count and name, and the top 25 became candidate slots, each named, defined, and given two synthetic examples by \texttt{deepseek-v4-pro}. The other 357 clusters, holding 780 of the 1,360 codes (57\%), never reached review. Most occur in a single report (209 clusters) or fragment concepts that the final slots cover. But ranking each cluster by its own report count also penalised concepts that the coder named inconsistently across separately coded reports, a redundancy \citet{wang-etal-2025-lata} also observed: protection and the hedging of unconfirmed figures were split across several smaller clusters, and the largest logistics cluster, like a revision-language cluster, reached 12 reports but lost an alphabetical tie-break with the last two candidates. The sector-needs, sector-response, and access slots can absorb some logistics and protection content, but no slot records hedging.

\paragraph{Human adjudication and audit trail.}
The author reviewed every candidate, merged overlapping concepts, renamed categories, rewrote definitions, and supplied two synthetic examples per slot. The frozen schema records the approver and approval time. The audit trail retains sampled report identifiers, per-report codes, and consolidation decisions. The final slots are: data currency and report metadata; sourcing and attribution; casualties and affected population; displacement and accommodation sites; health, disease outbreak and nutrition; physical, economic and agricultural damage; funding and donor contributions; education; WASH; food security and assistance; access constraints; coordination arrangements; sector needs and gaps; sector response actions; and shelter and non-food items.

\section{Measurement}
\label{sec:measurement}
\paragraph{Operational completeness.}
Let $j$ index the judge, $d$ a physical document, $z\in\mathcal{Z}$ one of 15 slots, and $s_j(d,z)\in\{0,.5,1\}$ the score for \textsc{absent}, \textsc{partial}, or \textsc{present}. Let $\mathcal{O}_j(d)\subseteq\mathcal{Z}$ be the slots with a parsable judgement and $m_j(d)=|\mathcal{O}_j(d)|$. Operational completeness is
\begin{equation}
C_j(d)=\frac{1}{m_j(d)}\sum_{z\in\mathcal{O}_j(d)}s_j(d,z).
\label{eq:completeness}
\end{equation}
Thus a missing judgement is not scored zero or imputed: it is dropped and reduces the denominator; a document with $m_j(d)=0$ is excluded. Each decision must include an evidence span. For event $e$, exact generator/prompt arm $a$, and $D_{j,a,e,z}$, the unique documents in that arm with a usable judgement for $z$, slot coverage is
\[
\begin{aligned}
\operatorname{coverage}_j(z;a,e)
&=\frac{1}{|D_{j,a,e,z}|}\\[-2pt]
&\quad\sum_{d\in D_{j,a,e,z}}s_j(d,z).
\end{aligned}
\]
Humans form one arm; the four machine arms remain separate. We do not pool models or prompts when locating a gap: $M_j(z,e)=\max_{a\in\mathcal{A}_{\rm machine}}\operatorname{coverage}_j(z;a,e)$ and $G_j(z,e)=\operatorname{coverage}_j(z;\mathrm{human},e)-M_j(z,e)$. The thresholded binary flag invoked in Recommendation 4 is explicitly
\[
B_j(z,e)=\mathbf{1}\!\left[
\begin{array}{l}
\operatorname{coverage}_j(z;\mathrm{human},e)\geq .8,\\[-2pt]
M_j(z,e)\leq .6
\end{array}\right].
\]
The \emph{human ceiling} is the mean of $C_j(d)$ over an event's evaluated professional reports, not an assumed 1.0. For Idai these are the 56 reports from which the schema was induced, so the Idai ceiling is in-sample and may favour human reports; Irma, Maria, Puebla, and Kerala are held out from schema induction. Per-judge, per-arm tables release every numerator, denominator, and score sum.

Two judges, \texttt{openai/gpt-5.6-luna} and \texttt{google/gemini-3.7-flash}, score all five events. Neither shares a model family with the DeepSeek generator models. Idai additionally retains \texttt{deepseek-v4-pro}, which does share the generator family, and \texttt{qwen/qwen3.8-flash}, which does not, yielding 12 event--judge cells. Figure~\ref{fig:completeness} uses 1,000 within-arm resamples of unique document scores, a percentile 95\% interval, and \texttt{random.Random(17)}. The arm-minus-human contrast uses 10,000 independent two-sample bootstrap resamples and a 100,000-draw, two-sided label-permutation test (seed 17); reports are not naturally paired across arms.

\paragraph{Content gate and judge reliability.}
The original prompt requires substantive content, but we additionally re-score schema-guided \textsc{present} cells deterministically. A matched heading with an empty or explicitly no-information body is capped at \textsc{absent}; a non-negated body with fewer than eight alphabetic tokens is capped at \textsc{partial}; other scores are unchanged.

For human validation, a seed-17 slot-stratified Idai sample contains 100 unique document--slot items, balanced between professional and machine reports and spanning all 15 slots. The author applied the same three-category rubric while blind to the withheld verdicts of all four Idai judges. For each judge, we report exact-match accuracy, unweighted nominal Cohen's $\kappa$, and linear-weighted $\kappa$ as an ordinal sensitivity check; for primary Gemini we additionally report one-vs-rest precision, recall, and F1 for each label, treating the human label as ground truth. A missing model verdict is dropped only for that judge and reduces its stated denominator. Our project plan, written before annotation, set the acceptance rule as judge agreement $\geq.7$ on a 100-item hand-scored sample without naming the statistic; our pre-annotation scoring code applied it to raw agreement, but some pre-annotation notes restated it as Cohen's $\kappa\geq.7$, so we report both. The Luna--Gemini Fleiss values are a separate inter-judge consistency check outside the acceptance rule, computed on unique complete-case document--slot items with nominal categories and two fixed raters per event.

\paragraph{Figure grounding and source availability.}
A regex extractor identifies numeric values in machine reports, same-day human reports, and the complete same-day social stream. Figure precision is the share of machine values also in the human report; recall reverses the denominator; casualty recall restricts the comparison to deaths, injuries, and missing people. A source-availability test asks both whether machine figures are traceable to the stream and whether official figures occur anywhere in that stream. Official-in-stream is averaged over unique valid event-days because its value repeats across machine arms; other cells average valid same-day machine--official pairs. Undefined denominators are dropped. Percentile 95\% CIs use 10,000 seed-17 resamples of event-days for official-in-stream and pairs for the other cells. A seed-17 manual audit samples 50 official figures marked stream-absent plus 25 official--stream and 25 machine--stream exact matches; two-sided Clopper--Pearson intervals bound each error rate. This analysis contains no model judge.

\paragraph{Metric meta-evaluation.}
We score each non-empty machine report against same-day or nearest same-window human references using ROUGE-L, BERTScore, and a schema-blind LLM quality score produced by \texttt{google/gemini-3.7-flash}, the same model used as the Gemini slot judge. Meta-evaluation includes only the three primary events, Idai, Irma, and Maria, because each has a substantial reference corpus; Puebla and Kerala each have only eight same-day pairs and weaker or genre-contaminated references. We compute per-event Spearman correlations with operational completeness. Reported $p$-values are SciPy 1.18's two-sided asymptotic \texttt{spearmanr} values; Table~\ref{tab:corr} applies Holm--Bonferroni correction jointly across its nine tests. The quality judge sees only a request to rate usefulness, clarity, organisation, and credibility on a 1--10 scale. Appendix~\ref{app:repro} specifies the exact implementations and prompts.

\section{Results}

\subsection{Presence produces a nominal inversion}

\begin{figure}[t]
\centering
\includegraphics[width=\columnwidth]{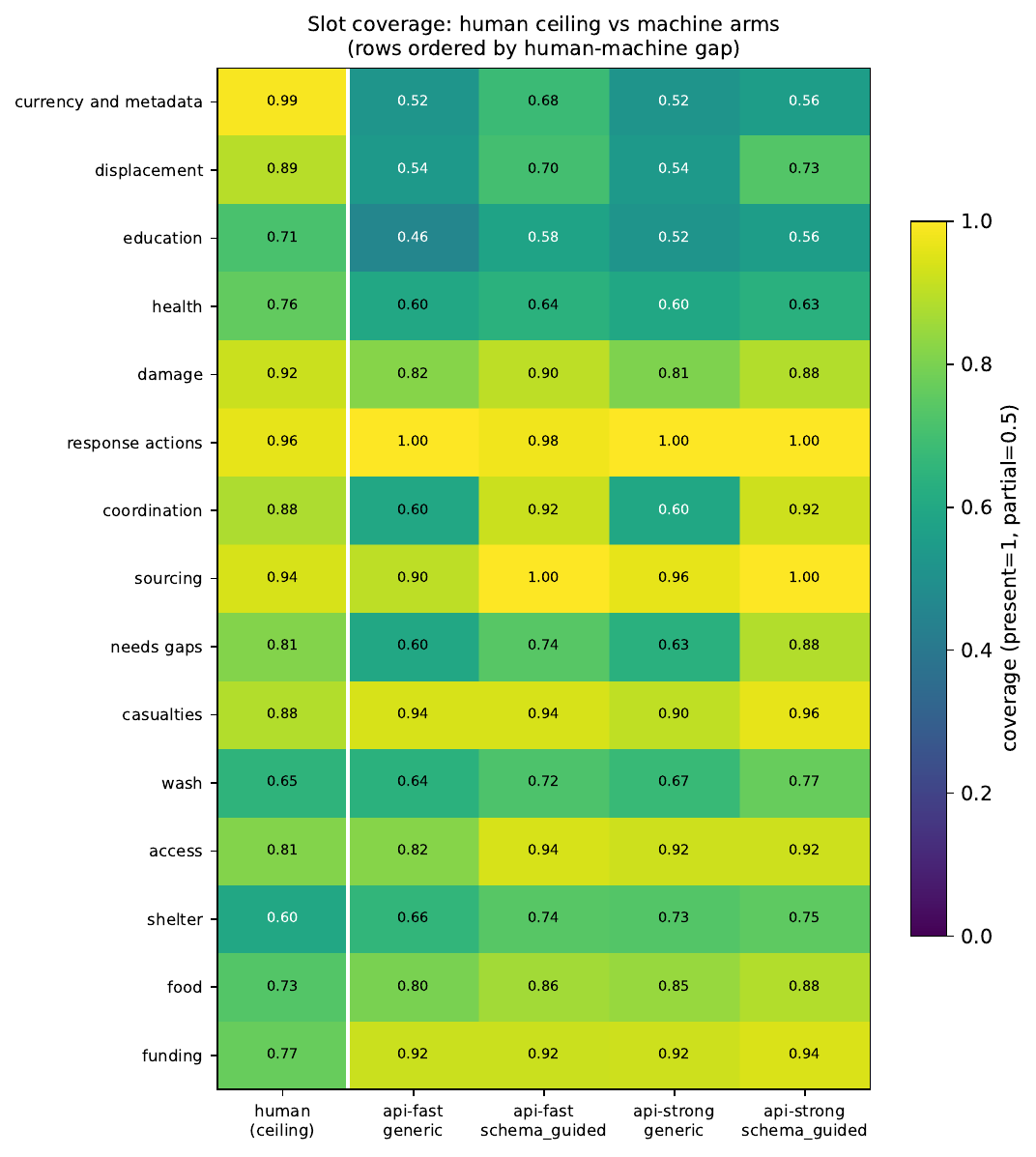}
\caption{Idai slot coverage under judge \texttt{google/gemini-3.7-flash}. Rows are ordered by $G_j(z,e)$, human coverage minus the best exact machine arm.}
\label{fig:coverage}
\end{figure}

\begin{figure}[t]
\centering
\includegraphics[width=\columnwidth]{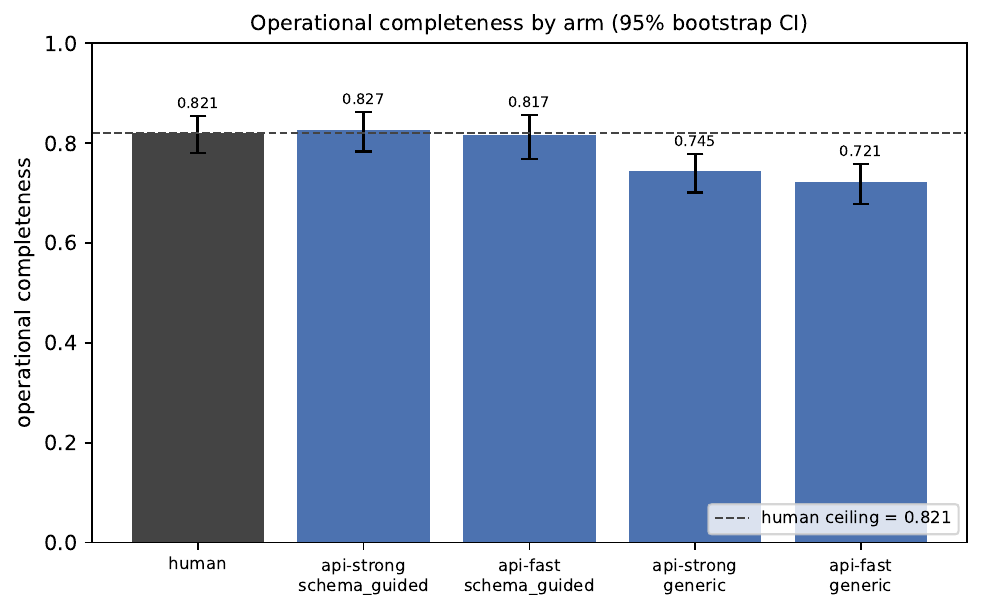}
\caption{Idai $C_j(d)$ under judge \texttt{google/gemini-3.7-flash}; percentile 95\% CIs resample unique documents within each arm (1,000 draws, seed 17). The nominal machine--human difference is not significant.}
\label{fig:completeness}
\end{figure}

Figure~\ref{fig:coverage} shows the central mismatch. Under \texttt{google/gemini-3.7-flash} on Idai, machine arms address most schema slots as often as or more often than human reports. In Figure~\ref{fig:completeness}, \texttt{api-strong/schema-guided} is nominally above the empirical human ceiling: 0.827 [.783,.862] versus 0.821 [.781,.854]. This $+.0061$ difference is statistically indistinguishable from zero (95\% difference CI $[-.0490,.0591]$; permutation $p=.852$).

The content-gated score remains nominally above under Gemini (0.824 vs. 0.821), so heading-only credit does not fully explain that judge's ordering; it remains statistically unsupported. The ordering is single-judge: under \texttt{openai/gpt-5.6-luna}, the best Idai machine arm scores 0.819 versus a human ceiling of 0.845, and the DeepSeek and Qwen judges also place the human ceiling above every machine arm. It is also genre-dependent: without Idai's three snapshots, the Gemini human ceiling rises to 0.834, above every machine arm. As a supplementary small-corpus observation rather than replication of Idai, the descriptive mean over Kerala's 20 machine reports is 0.800 versus 0.743 for seven humans under Luna; its four arms are pooled only for this observation, not for gap or arm comparisons.

\subsection{One gap is structural}

Data currency and report metadata has $G_j(z,e)>0$ in all 12 observed event--judge cells, spanning five events, three hazard types (tropical cyclone, earthquake, flood), and four judge families. Applying Table~\ref{tab:corpus}'s manifest-keyed document rule gives a range of +0.200 to +0.458. The conclusion is unchanged when the 159-loss judge is excluded: removing Qwen's entire cell leaves every remaining cell positive (11/11). Under Qwen's observed-case analysis, 28/56 human versus 3/102 machine judgements are missing for this slot, so we do not use that cell to claim robustness to missingness.

The result separates medium-driven from event-driven gaps. Under \texttt{openai/gpt-5.6-luna}, Idai displacement looks like a second ceiling slot ($G_j=+0.288$), but the same defined quantity changes across Idai, Irma, and Maria (+0.288, $-$0.035, +0.070). Our a priori medium argument was that a social post is not the issuing instrument for a sitrep's organisation, report number, publication date, or stated authority. Empirically, stream-derived arms supplied that provenance less often than humans in all 12 observed cells. Displacement instead depends on what an emergency makes salient in public communication. A single-event study would have published it as a ceiling slot. The absolute flag $B_j$ is less stable: for Idai metadata, $M_j=.520/.540/.615/.680$ across DeepSeek/Luna/Qwen/Gemini, so the .6 boundary splits the judges even though all gaps are positive.

\subsection{The ceiling is in the source}

\begin{table*}[t]
\centering
\scriptsize
\setlength{\tabcolsep}{2.3pt}
\begin{tabular}{lrrrrrr}
\toprule
Event & Pairs & Official in stream & Machine in stream & Precision & Recall & Casualty recall \\
\midrule
Idai & 83 & 14.3 [9.5,19.7] & 85.4 [83.7,87.3] & 14.9 [12.9,17.0] & 11.0 [8.7,13.5] & 33.0 [25.7,40.7] \\
Irma & 28 & 14.3 [7.0,23.1] & 90.1 [87.4,92.8] & 16.3 [12.4,20.1] & 5.4 [4.0,6.8] & 52.1 [40.6,62.5] \\
Maria & 20 & 19.6 [11.6,28.3] & 85.8 [81.8,90.0] & 15.3 [10.2,21.0] & 8.9 [6.4,11.8] & 12.5 [0.0,25.0] \\
\emph{Puebla} & \emph{8} & \emph{36.2 [35.3,37.0]} & \emph{93.2 [90.4,96.1]} & \emph{18.9 [16.4,21.9]} & \emph{21.0 [17.3,24.3]} & \emph{36.9 [29.3,44.5]} \\
\emph{Kerala} & \emph{8} & \emph{39.0 [25.0,52.9]} & \emph{90.7 [89.1,92.4]} & \emph{5.9 [3.9,7.4]} & \emph{21.9 [14.3,28.1]} & \emph{22.9 [14.6,31.2]} \\
\bottomrule
\end{tabular}
\caption{Regex grounding percentages with percentile-bootstrap 95\% CIs (10,000 draws, seed 17); no model judge. ``Official'' and ``machine'' mean exact-number matches to the day's stream. Official-in-stream averages and resamples unique valid event-days ($n=21/7/4/2/2$); all other cells average and resample valid same-day machine--official pairs. Undefined denominators are dropped: casualty-recall $n=79/8/12$ for Idai/Irma/Maria and Maria recall $n=16$; other non-official cells use Pairs. Italicised events have only eight pairs. In manual audit, 5/50 declared absences were errors (10.0\%, exact 95\% CI [3.3,21.8]), as were 21/25 official matches (84.0\% [63.9,95.5]) and 1/25 machine matches (4.0\% [0.1,20.4]).}
\label{tab:source}
\end{table*}

Table~\ref{tab:source} resolves the paradox without a model in the loop, while also exposing the regex's limits. Across Idai, Irma, and Maria, exact-number recall is 11.0\%, 5.4\%, and 8.9\%; casualty recall is 33.0\%, 52.1\%, and 12.5\%. Raw matching links 14.3--19.6\% of official figures and 85.4--90.1\% of machine figures to the stream. Manual review confirms 45/50 declared official absences, but only 4/25 official matches are semantic, versus 24/25 machine matches. Thus the official column is an ungated exact-match rate, inflated by common-number collisions, while sampled machine traceability is substantially more reliable. Puebla and Kerala are not averaged into the headline: each has only eight same-day pairs, Puebla's references are figure-sparse short products, and Kerala has the densest stream in the corpus.

\subsection{Standard metrics fail differently}

\begin{table}[t]
\centering
\scriptsize
\setlength{\tabcolsep}{1pt}
\begin{tabular}{lrrr}
\toprule
Metric & Idai & Irma & Maria \\
\midrule
ROUGE-L & +.070 (.486) & +.178 (.483) & $-$.272 (.310) \\
BERTScore & $-.426^{***}$ (.0001) & $-.469^{**}$ (.006) & $-.439^{*}$ (.023) \\
Generic judge & +.250 (.067) & $-$.205 (.483) & $-$.375 (.067) \\
\bottomrule
\end{tabular}
\caption{Spearman $\rho$ with $C_j(d)$ ($n=102/48/43$); parentheses are Holm-adjusted $p$ over all nine tests. $^{*}p<.05$, $^{**}p<.01$, $^{***}p<.001$. Completeness judge: \texttt{google/gemini-3.7-flash} for Idai and \texttt{openai/gpt-5.6-luna} for Irma/Maria. The schema-blind quality judge in every ``Generic judge'' cell is \texttt{google/gemini-3.7-flash}.}
\label{tab:corr}
\end{table}

\begin{figure*}[t]
\centering
\includegraphics[width=0.98\textwidth]{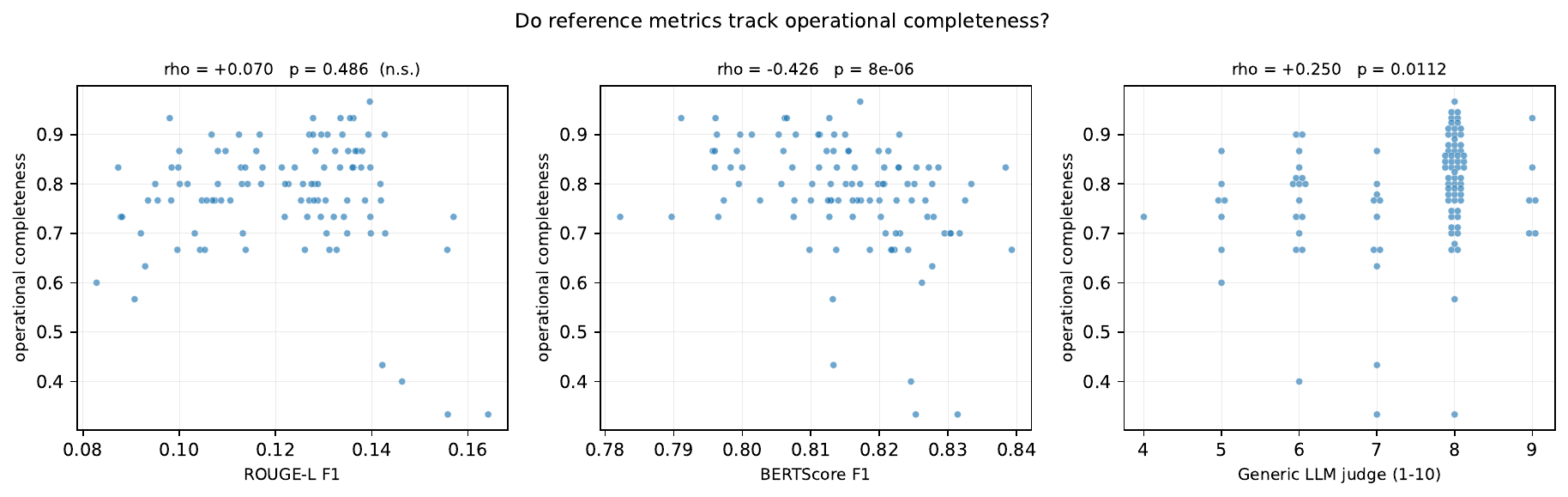}
\caption{Idai reference metrics against $C_j(d)$ scored by \texttt{google/gemini-3.7-flash}; the schema-blind quality score is from the same model. Panel $p$-values are raw; Table~\ref{tab:corr} gives the nine-test Holm correction. In the generic-judge panel, exactly coincident points are spread slightly sideways and upward for visibility; statistics use the unmodified values.}
\label{fig:corr}
\end{figure*}

Table~\ref{tab:corr} and Figure~\ref{fig:corr} show three distinct failures on the three primary events. ROUGE-L is not significant on any of them and changes sign. BERTScore is significantly negative on all three primary events after Holm correction, with a stable magnitude of $-0.43$ to $-0.47$. This association is partly length-mediated: completeness correlates +0.532 with length and BERTScore correlates $-0.446$ on Idai; the partial correlation controlling for length remains $-0.248$. The metric penalises some of the elaboration that slot completeness rewards.

The generic LLM judge is less stable still: its correlation is positive on Idai and negative on Irma and Maria; neither formerly nominally significant cell survives correction. It is also circular within the pilot. When the generic judge and completeness scorer share a model family, $\rho=+0.250$ (raw $p=.011$); across families, $\rho=+0.174$ (raw $p=.080$). An evaluator selected on one disaster or model family can therefore transfer poorly.

To match prior system-level critiques at their level of analysis, we also average reports within each of the four generator--prompt arms and correlate the four arm means, descriptively only. Across Idai/Irma/Maria, BERTScore's arm-level $\rho$ is $-1.0/-1.0/-.8$, ROUGE-L's is $-.8/.0/-.8$, and the Gemini quality judge's is $+.4/-.6/-.6$. With only four systems per event these are order summaries, not inferential evidence, but the BERTScore reversal is not solely a report-level phenomenon.

\subsection{Human validation is judge-specific}

\begin{table}[t]
\centering
\scriptsize
\setlength{\tabcolsep}{3.8pt}
\begin{tabular}{lrrrr}
\toprule
Judge & $n$ & Accuracy & $\kappa$ & $\kappa_{\rm lin}$ \\
\midrule
DeepSeek & 100 & .790 & .592 & .627 \\
Luna & 99 & .828 & .617 & .668 \\
Qwen & 90 & .856 & .681 & .715 \\
Gemini & 100 & .910 & .799 & .806 \\
\addlinespace[2pt]
\multicolumn{5}{l}{\emph{Primary Gemini by human label}} \\
\cmidrule(lr){1-5}
Label & Support & Precision & Recall & F1 \\
\midrule
\textsc{absent} & 9 & .875 & .778 & .824 \\
\textsc{partial} & 19 & .773 & .895 & .829 \\
\textsc{present} & 72 & .957 & .931 & .944 \\
\bottomrule
\end{tabular}
\caption{Blind author agreement on 100 seed-17, slot-stratified Idai document--slot items (50 professional, 50 machine). Accuracy is exact label agreement; $\kappa$ is unweighted nominal Cohen's $\kappa$; $\kappa_{\rm lin}$ is a sensitivity check. The project plan set agreement $\geq.7$ without naming the statistic (Section~\ref{sec:measurement}). Gemini class metrics treat human labels as ground truth. Missing verdicts reduce $n$ (Luna: 1; Qwen: 10). Judge models are \texttt{deepseek-v4-pro}, \texttt{openai/gpt-5.6-luna}, \texttt{qwen/qwen3.8-flash}, and \texttt{google/gemini-3.7-flash}.}
\label{tab:human-validation}
\end{table}

All four judges reach the plan's .7 threshold on raw agreement, the statistic our pre-annotation scoring code applied. Only Gemini also reaches .7 on the separately reported chance-corrected $\kappa$; it is the judge behind the Idai nominal inversion and Figures~\ref{fig:coverage}--\ref{fig:completeness}. Its agreement is high on both professional-report items ($\kappa=.861$, raw .940) and machine-report items ($\kappa=.730$, raw .880). Luna, which supplies the principal cross-event check, falls from $\kappa=.762$ on professional-report items to .470 on machine-report items. The labels therefore support the Gemini-based pilot measurement more strongly than the Luna-based replication, rather than validating LLM judging in general.

\begin{table}[t]
\centering
\scriptsize
\setlength{\tabcolsep}{3.5pt}
\begin{tabular}{lrrr}
\toprule
Event & $n_H$ & Luna human & Gemini human \\
\midrule
Idai & 56 & .845 [.808,.875] & .821 [.781,.854] \\
Irma & 19 & .739 [.695,.777] & .709 [.653,.758] \\
Maria & 25 & .763 [.696,.825] & .717 [.636,.789] \\
\emph{Puebla}$^\dagger$ & 14 & .567 [.493,.640] & .500 [.431,.564] \\
\emph{Kerala} & 7 & .743 [.695,.781] & .779 [.700,.839] \\
\bottomrule
\end{tabular}
\caption{Per-event empirical human ceiling, the mean of $C_j(d)$ with 95\% document-bootstrap CIs. Judges are \texttt{openai/gpt-5.6-luna} and \texttt{google/gemini-3.7-flash}. $^\dagger$Puebla is the collected-product score, not used as a sitrep ceiling because 10/14 products are short infographics or snapshots.}
\label{tab:human-ceilings}
\end{table}

\section{Implications for Humanitarian Practice}
\label{sec:practice}

We ran no deployment or practitioner study, and our drafts came from two DeepSeek models reading English-language posts, so what follows is what the measurements imply for that setting, not an observed effect; nor did we measure whether drafts add value by arriving before official reports.

\paragraph{For information managers and report authors.}
On Idai, under our best-validated judge, drafts prompted with the schema covered it about as fully as professional reports (Figure~\ref{fig:completeness}), although truncating long professional reports for judging understates their coverage. Their figures and provenance were the weak point: on Idai, Irma, and Maria, exact matching linked only 5--11\% of same-day official figures to them (Table~\ref{tab:source}), and on every event they stated their own as-of date, issuing organisation, and update number less often than professional reports. A draft can therefore look finished without being sourced: reviewers should check its figures, dates, and sources, not its headings. It is best treated as an unverified situational signal whose figures need re-sourcing from official or assessment channels and whose provenance must come from the issuing office; it should not circulate under an agency's name. Because generations can repeat phone numbers and email addresses from crisis posts (Appendix~\ref{app:repro}), drafts also need screening for personal data.

\paragraph{For organisations piloting or funding reporting tools.}
Similarity to professional reports and generic LLM quality ratings did not track completeness here (Table~\ref{tab:corr}). Pilots and procurement can instead test a tool on past emergencies: check its figures against same-day official reports, compare it with professional reports of the same type rather than with a perfect score, and ask which inputs it used (Section~\ref{sec:shared-task}).

\paragraph{For tool builders.}
Most official figures never appeared in the day's HumAID posts, a labelled sample of English-language tweets, so a stronger model could not have recovered them from that input. Drafting is therefore likely to be more useful when connected to the channels official figures come from, with social media flagging emerging needs and locations for verification; we did not test this. Gaps also vary by event: drafts covered displacement much less often than professional reports on Idai, but not on Irma or Maria, so a gap seen in one emergency needs re-checking in the next.

\paragraph{For professional reporting itself.}
Professional reports also scored below 1.0 (Table~\ref{tab:human-ceilings}), partly because shorter products such as snapshots and infographics are not meant to cover every slot. A practitioner-reviewed slot list, adapted to each product type and extended with categories ours lacks or only partly covers, such as protection, could help authors check for unintended omissions; whether that improves operational use needs evaluation with practitioners.

\section{What the Shared-Task Community Should Adopt}
\label{sec:shared-task}
As of 31 August 2026, the official LT4CPR guidelines score report sections 3--11 using text similarity, bullet-level alignment, and confidence-label agreement, then average replicates within windows, windows within crises, and crises; the public task page also named ROUGE, BLEU, and BERTScore as anticipated automatic metrics. The official results, released since, rank systems by the mean of BERTScore F1 and BLEURT \citep{lt4cpr-2026-shared-task}; because BERTScore was negatively associated with completeness on all three primary events (Table~\ref{tab:corr}), that ranking need not track operational completeness, although we did not test BLEURT or the task's own reference reports. The results suggest seven explicit deltas to that protocol rather than another single metric.

\begin{enumerate}
\item \textbf{Use an empirical human ceiling.} Across the three primary events and two judges in Table~\ref{tab:human-ceilings}, human reports score 0.71--0.85, so a system score is not a percentage of perfection. \emph{Delta:} add human calibration, which the current protocol does not define.
\item \textbf{Do not rank systems by presence alone.} Slot coverage can place machine reports nominally above professional ones without a significant difference or cross-judge replication. \emph{Delta:} show bullet-level alignment (now only a secondary workbook metric) and confidence-label agreement (specified in the guidelines but not computed by the official evaluator) beside the similarity-based primary score on the leaderboard, and replace any presence-only leaderboard interpretation.
\item \textbf{Require figure-level grounding.} Report precision, recall, and casualty recall against the official record alongside slot coverage; Table~\ref{tab:source} gives primary-event casualty recall of 33.0/52.1/12.5\%. \emph{Delta:} add number-specific exactness beyond text similarity and bullet alignment.
\item \textbf{Report $G_j$ with intervals, not $B_j$.} The 0.6 inequality in Section~\ref{sec:measurement} lies inside the observed .520/.540/.615/.680 spread for one slot; two judges flag it and two do not. \emph{Delta:} add uncertainty to the current point-score aggregation.
\item \textbf{Report per-judge missingness.} Unparsable outputs were 0/19/159/0 across four judges, with the largest loss concentrated in the slot carrying the largest gap. \emph{Delta:} keep the evaluator's coverage and metric-status checks, and extend them to per-judge parse-failure rates whenever an LLM judge is used.
\item \textbf{Use a human-validated cross-family judge, or disclose that this was not done.} Same-family agreement can masquerade as validity, and Table~\ref{tab:human-validation} shows that cross-family judges are not interchangeable. \emph{Delta:} add blind human validation and cross-model robustness when an LLM judge supplements official metrics.
\item \textbf{State the source ceiling.} When inputs are social media, evaluation must quantify what share of reference facts the input actually contains. \emph{Delta:} report the task's existing input-to-reference audit with the official results, and extend it from the tweets each reference cites to the whole input, distinct from output-to-reference scoring.
\end{enumerate}

Together these measures separate three questions that a shared task should not collapse: whether a report names the right topics, whether its claims match an operational reference, and whether the supplied evidence could support those claims.

\section{Conclusion}

Professional reporting practice reveals a sharp distinction between looking complete and being operationally grounded. Under one judge, an LLM arm is nominally above but statistically indistinguishable from the empirical human ceiling; the ordering does not replicate under Luna and is smaller than the truncation bias measured among sampled affected reports. Across the measured primary events, systems reproduced little of the official quantitative record, and across all 12 observed event--judge cells they supplied issuing provenance less often than professionals. Crisis-report evaluation must measure source availability, figure grounding, and judge-indexed human-relative completeness together. For practitioners, a draft built from social media largely restates figures already in that day's posts; it cannot stand in for the official record or carry an agency's authority.

\section*{Limitations}
\paragraph{Judge validity.}
The blind validation in Table~\ref{tab:human-validation} uses one annotator, who also approved the schema, and only Idai; it may therefore overstate agreement with independent practitioners and does not validate transfer to other events. All four judges reach the plan's .7 threshold on raw agreement, but only Gemini reaches .7 on the separately reported chance-corrected $\kappa$. On unique document--slot complete cases, unweighted nominal Fleiss' $\kappa$ between Luna and Gemini is 0.653 ($n=2{,}351$), 0.646 ($n=1{,}005$), and 0.707 ($n=1{,}020$) on Idai, Irma, and Maria. These values lie outside the acceptance rule and measure consistency, not correctness; Qwen's missingness is non-random. All LLM-judged conclusions should therefore be read alongside the model-independent source-availability result.

\paragraph{Schema and corpus scope.}
The schema was induced from all 56 Idai reports, 44 of them OCHA's, so it chiefly reflects OCHA reporting practice, and its Idai ceiling is in-sample; held-out professional reports on Irma and Maria score 0.71--0.76 (Table~\ref{tab:human-ceilings}), below Idai's 0.82--0.85. Its conceptual space was shaped before human review (Section~\ref{sec:schema}): our instructions seeded five reporting conventions, one model proposed the codes from truncated text for the 12 longest reports, a 25-candidate cap left clusters holding 57\% of the codes unreviewed, with drafted consolidation, capped in advance at 15 slots. The seeded provenance conventions are nonetheless pervasive in practice: an LLM-free lexicon released with our code counts 395 as-of date markers and 413 attribution phrases in the same reports. The model that proposed the codes and named the candidates, \texttt{deepseek-v4-pro}, is also the \texttt{api-strong} generator, whose schema-guided arm receives the slot names and definitions; slots shaped by its salience could favour its outputs. Completeness therefore describes the 15 measured slots, not all of reporting practice, and the omissions cut both ways: agency-reported categories such as protection and logistics would likely favour professional reports, whereas public-facing ones such as search and rescue or hazard forecasts could favour stream-derived drafts. No second researcher or practitioner independently coded the reports, so schema construction has no inter-coder reliability estimate; the released audit trail allows others to run that check. These limits bear on the completeness-based results (the inversion, the provenance gap, the human ceilings, and the metric correlations), not on figure grounding or source availability, which use neither the schema nor a model judge. HumAID and the generated reports are English-language, all drafts come from one generator family, and HumAID is itself a labelled sample of about 77K tweets drawn from about 24M \citep{alam-etal-2021-humaid}, so the source ceiling describes this sample rather than social media at large; the ReliefWeb collection includes a small number of non-English products, limiting multilingual claims. Irma and Maria have 19 and 25 evaluated human reports, and Kerala has seven; their ceiling estimates are noisier than Idai's. Puebla is genre-contaminated: 10 of 14 judged documents are infographics or snapshots, lowering its Luna collected-product score from 0.725 for sitreps alone to 0.567. We therefore treat Puebla and Kerala as supplementary.

\paragraph{Measurement scope.}
Regex figure matching can count coincidental numeric matches and miss semantically equivalent expressions. The manual audit quantifies both modes: 5/50 declared absences and 21/25 official exact matches were erroneous, while only 1/25 machine exact matches was erroneous. A machine figure absent from an official report is unverified, not necessarily false. Original judge inputs truncated reports after 20,000 characters; 56/100 evaluated primary-event professional reports exceeded that cap (30/56 Idai, 11/19 Irma, 15/25 Maria). We re-judged a seed-17 random sample of 12 affected reports (four per primary event) at full length with Gemini: the paired ceiling shift was +0.056 [+.031,+.081] (10,000 document-bootstrap draws). This quantifies bias among long reports, not the whole corpus, but is larger than the nominal Idai machine advantage. Presence also risks rubric leakage; our deterministic body gate lowered the nominal Gemini machine score from .827 to .824 but did not reverse it.

\section*{Ethical Considerations}
\paragraph{Data protection:} ReliefWeb reports are public. Social-media text is not redistributed, no user identifiers appear, and no post is quoted verbatim.
\paragraph{Dual use and deployment hazards:} The system is intended as evaluation infrastructure, not an autonomous reporting tool. An operationally incomplete report that appears complete could misdirect scarce resources or falsely convey authority; the statistically unsupported nominal ordering still illustrates this deployment hazard.

\bibliography{references}

\appendix
\section{Reproducibility and Audit Trail}
\label{app:repro}

All LLM calls pass through one client with a disk cache, a fail-closed cost gate, and a JSONL log containing request hashes, token counts, costs, and lengths but never text. Seeds are fixed at 17. The recorded spend is \$31.35, including \$1.72 for the untruncated sensitivity run. The code and data release is available at
\url{https://github.com/techsachinkr/llm-sitrep-evaluation/}.
It includes \texttt{supplementary/reproducibility/}: every prompt verbatim with sentinel payloads, the frozen 15-slot schema with definitions and two synthetic examples per slot, public model identifiers and call dates, the 125-report ReliefWeb manifest, and text-free matcher-audit labels with context hashes. The repository also releases the schema audit trail (\texttt{data/processed/schema/}): the induction sample, every open code's name and definition by report, all 382 clusters, the 25 candidates, and the merge rationales, without evidence spans. The manifest specifies first-20,000-Python-character truncation followed by \texttt{[...truncated...]}, as well as titles, sources, verified dates, retrieval locators, hashes, and exclusions.

\begin{table*}[t]
\centering
\scriptsize
\setlength{\tabcolsep}{3pt}
\begin{tabular}{lllll}
\toprule
Role & Public model identifier(s) & Temp./top-$p$ & Before fix & Corrected cap \\
\midrule
Open coding & \texttt{deepseek-v4-pro} & default / unsent & 6,000 requested, unsent & 6,000 \\
Cluster naming & \texttt{deepseek-v4-pro} & default / unsent & 900 requested, unsent & 900 \\
Generation & \texttt{deepseek-v4-pro}, \texttt{deepseek-v4-flash} & default / unsent & 2,000 requested, unsent & 48,000 \\
Idai slot judging & \texttt{deepseek-v4-pro}, \texttt{openai/gpt-5.6-luna} & default / unsent & DS 400 unsent; Luna 400 & 400 \\
Idai slot judging & \texttt{qwen/qwen3.8-flash}, \texttt{google/gemini-3.7-flash} & default / unsent & 3,500; 1,200 & unchanged \\
Replication slot judging & \texttt{openai/gpt-5.6-luna}, \texttt{google/gemini-3.7-flash} & default / unsent & 1,200 or 3,500 & unchanged \\
Generic quality & \texttt{google/gemini-3.7-flash} & default / unsent & 512 & unchanged \\
Full-text sensitivity & \texttt{google/gemini-3.7-flash} & default / unsent & 1,200 & unchanged \\
\bottomrule
\end{tabular}
\caption{Decoding parameters by LLM role. Temperature was omitted (provider default); the wrapper does not implement or transmit top-$p$. ``Unsent'' is the first defect listed at the end of this appendix: the DeepSeek endpoint ignored the wrong parameter name. The Idai generation, coding, naming, and DeepSeek-judge calls are the pre-fix reported calls; four-event replication generation used the corrected 48,000 cap. The artifact disaggregates every event--judge cap and records effort.}
\label{tab:decoding}
\end{table*}

The publication tables are generated by \texttt{python -m src.paper\_analysis}. \texttt{corpus\_manifest.csv} records document and judgement denominators; \texttt{slot\_coverage\_by\_judge\_arm.csv} releases every $\operatorname{coverage}_j$ numerator and denominator; the document, inversion, content-gate, truncation, agreement, Holm, and four-arm tables reproduce the remaining claims. \texttt{human\_validation\_pairs.csv} releases the 100 text-free human and withheld judge labels, and \texttt{human\_validation\_summary.csv} recomputes Table~\ref{tab:human-validation}, including per-judge exclusions. \texttt{per\_judgement.csv} releases the text-free raw rows, including failures and duplicate keys, so the last-row/observed-case rules are independently checkable. Figure~\ref{fig:completeness} reads the corrected Gemini table directly.

\paragraph{Measurement specification.}
The artifact publishes the exact number and date-exclusion regexes, scale dictionary, unit handling, and matching rule: remove comma, ASCII space, and non-breaking space; apply thousand/million/billion/lakh/crore scales; keep percentages separate; deduplicate within documents; drop recognised dates and bare years 1900--2100; and match \texttt{(unit, round(value,6))} exactly with no tolerance. It pins \texttt{rouge-score==0.1.2} with \texttt{RougeScorer(['rougeL'], use\_stemmer=True)} and maximum-F1 multi-reference selection; \texttt{bert-score==0.3.13}, \texttt{roberta-large} revision \texttt{722cf37b}, English, no baseline rescaling, and multi-reference maximum; and \texttt{sentence-transformers/all-MiniLM-L6-v2} revision \texttt{1110a243}, normalised embeddings, average-linkage precomputed cosine distance, and threshold .35. The full revisions, tokenisation, packages, runtime, and 76,484-row non-text HumAID ID/hash manifest are machine-readable.

Public machine-output text is deliberately omitted after a spot check found that generations can repeat phone numbers and email addresses from crisis posts. Each of 264 outputs instead has a request hash, output hash, call time, and public model ID; licensed HumAID records have IDs and text hashes. Thus the released numerical tables and measurement functions are bit-for-bit auditable, and acquired inputs can be hash-verified, but fresh generation from mutable aliases is not promised to be deterministic. This is a privacy and provider-version boundary, not an unreported omission.

Four silent implementation defects were found and corrected. First, the DeepSeek client sent \texttt{max\_completion\_tokens}, which that endpoint ignored; 75\% of 3,532 calls exceeded the configured cap. Second, hyphenated ReliefWeb keys and underscored HumAID keys prevented reference pairing; canonical normalisation repaired it. Third, a 32,000-token generation probe truncated 25\% of schema-guided versus 8\% of generic Irma map calls; the reported replication uses 48,000, leaving 1.0\% truncation overall and 0.0\% for schema-guided Irma. Fourth, reference-metric CSVs recorded the default judge alias even when \texttt{--judge-model judge\_gemini} was passed. Request logs establish \texttt{google/gemini-3.7-flash}; labels and code are corrected, with no numeric score changed. Regression tests cover all four repairs.

\end{document}